\documentclass[11pt]{article}

\usepackage{amsmath,amssymb,amsthm}
\usepackage{hyperref}
\usepackage[usenames]{color}
\usepackage[margin=1in]{geometry}
\usepackage{graphicx}
\usepackage{subfigure}
\newcommand{\todovk}[1]{}
\newcommand{\todoea}[1]{}
\newcommand{\eat}[1]{ }

\def\ie{{\it i.e.},~}
\def\eg{{\it e.g.},~}
\def\etal{{\it et al.}~}

\def\NP{\mathsf{NP}}
\def\RP{\mathsf{RP}}
\def\bnsel{\mathsf{BN\mbox{-}Sel}}
\def\optsel{\mathsf{Opt\mbox{-}Sel}}
\def\NZ{\mathsf{NZ}}
\def\sparsity{\mathrm{sparsity}}

\def\E{\mathbb{E}}

\def\reals{\mathbb{R}}

\def\th{{^{\textit{th}}}}

\def\corr{\mathrm{corr}}
\def\argmin{\mathrm{argmin}}

\def\Mut{\mathsf{Mut}}
\def\Neigh{\mathsf{Neigh}}
\def\naturals{\mathbb{N}}
\def\Bene{\mathsf{Bene}}
\def\Neut{\mathsf{Neut}}
\def\opt{\mathsf{opt}}
\def\best{\mathsf{Best}}
\def\evalg{{\mathcal EA}}
\def\Sel{\mathsf{Sel}}
\def\Dists{{\mathcal D}}
\def\loss{\mathrm{L}}
\def\lin{\mathsf{Lin}}

\newcommand{\lznorm}[1]{\mathrm{sparsity}(#1)}
\newcommand{\ltwonorm}[1]{\Vert #1 \Vert}
\newcommand{\ip}[2]{\langle #1, #2 \rangle}

\newtheorem{lemma}{Lemma}
\newtheorem{theorem}{Theorem}
\newtheorem{remark}{Remark}
\newtheorem{definition}{Definition}

\newtheorem{claim}{Claim}

\begin{document}
\title{Attribute-Efficient Evolvability of Linear Functions} 
\author{Elaine Angelino\thanks{This author is supported in part by a grant from
the National Library of Medicine (4R01LM010213-04) and NSF grant CCF-09-64401}\\
Harvard University \\ \texttt{elaine@eecs.harvard.edu} \and Varun
Kanade\thanks{This author is supported by a Simons Postdoctoral Fellowship.} \\
UC Berkeley \\ \texttt{vkanade@eecs.berkeley.edu}}

\maketitle

\begin{abstract}
In a seminal paper, Valiant (2006) introduced a computational model for
evolution to address the question of complexity that can arise through Darwinian
mechanisms. Valiant views evolution as a restricted form of computational
learning, where the goal is to \emph{evolve} a hypothesis that is close to the
\emph{ideal function}. Feldman (2008) showed that (correlational) statistical
query learning algorithms could be framed as evolutionary mechanisms in
Valiant's model. P. Valiant (2012) considered evolvability of real-valued
functions and also showed that weak-optimization algorithms that use
weak-evaluation oracles could be converted to evolutionary mechanisms.

In this work, we focus on the \emph{complexity} of representations of
evolutionary mechanisms. In general, the reductions of Feldman and P. Valiant
may result in intermediate representations that are arbitrarily complex
(polynomial-sized circuits). We argue that biological constraints often dictate
that the representations have low complexity, such as constant depth and fan-in
circuits. We give mechanisms for evolving sparse linear functions under a large
class of smooth distributions. These evolutionary algorithms are
attribute-efficient in the sense that the size of the representations and the
number of generations required depend only on the sparsity of the target
function and the accuracy parameter, but have no dependence on the total number of
attributes.

\end{abstract}

\newpage

\section{Introduction}
\label{sec:introduction}
Darwin's theory of evolution through natural selection has been a cornerstone of
biology for over a century and a half. Yet, a quantitative theory of complexity
that could arise through Darwinian mechanisms has remained virtually unexplored.
To address this question, Valiant introduced a computational model of
evolution~\cite{Valiant:2009-evolvability}.  In his model, an organism is an
entity that computes a function of its environment.  There is a (possibly
hypothetical) \emph{ideal function} indicating the best behavior in every
possible environment. The performance of the organism is measured by how close
the function it computes is to the ideal. An organism produces a set of
offspring, that may have mutations that  alter the function computed. The
performance (fitness) measure acting on a population of mutants forms the basis
of natural selection. The resources allowed are the most generous while
remaining feasible; the mutation mechanism may be any efficient randomized
Turing machine, and the function represented by the organism may be arbitrary as
long as it is computable by an efficient Turing machine.

Formulated this way, the question of evolvability can be asked in the language
of computational learning theory. For what classes of ideal functions, $C$, can
one expect to find an evolutionary mechanism that gets arbitrarily close to the
ideal, within feasible computational resources? Darwinian selection is
restrictive in the sense that the only feedback received is {\em aggregate} over life
experiences. Valiant observed that any feasible evolutionary mechanism could be
simulated in the statistical query framework of Kearns~\cite{Kearns:1998}. In a
remarkable result, Feldman showed that in fact, evolvable concept classes are
exactly captured by a restriction of Kearns' model, where the learning
algorithm is only allowed to make \emph{performance queries}, \ie it produces a
hypothesis and then makes a query to an oracle that returns the (approximate)
performance of that hypothesis under the
distribution~\cite{Feldman:2008-evolvability}.\footnote{Feldman calls these
correlational statistical queries, because when working with boolean
functions with range $\{-1, 1\}$, the performance of any hypothesis is its
correlation with the ideal function.} P.  Valiant studied the evolvability of
real-valued functions and showed that whenever the corresponding weak
optimization problem, \ie approximately minimizing the expected loss, can be
solved by using a weak evaluation oracle, such an algorithm can be converted
into an evolutionary mechanism~\cite{Valiant:2012-real}. This implies that a
large class of functions -- fixed-degree real polynomials -- can be evolved
with respect to any convex loss function.

Direct evolutionary mechanisms, not invoking the general reductions of Feldman
and P. Valiant, have been proposed for certain classes in restricted settings.
Valiant showed that the class of disjunctions is evolvable using a simple set of
mutations under the uniform distribution~\cite{Valiant:2009-evolvability}.
Kanade, Valiant and Vaughan proposed a simple mechanism for evolving homogeneous
linear separators under radially symmetric distributions~\cite{KVV:2010-drift}.
Feldman considered a model where the ideal function is boolean but the
representation can be real-valued, allowing for more detailed feedback. He
presents an algorithm for evolving large margin linear separators for a large
class of convex loss functions~\cite{Feldman:2011-LTF}. P. Valiant also showed
that with very simple mutations, the class of fixed-degree polynomials can be
evolved with respect to the squared loss~\cite{Valiant:2012-real}.

Current understanding of biology (or lack thereof) makes it
difficult to formalize a notion of \emph{naturalness} for mutations in these
frameworks; in particular, it is not well understood how mutations to DNA
relate to functional changes in an organism. That said, the more direct
algorithms are appealing due to the simplicity of their mutations.  Also, the
``chemical computers'' of organisms may be slow, and hence, representations that
have low complexity are attractive.  In general, Feldman's
generic reduction from statistical query algorithms may use arbitrarily complex
representations (polynomial-sized circuits), depending on the specific algorithm
used.  In the remainder of the introduction, we first describe a particular
class of biological circuits, \emph{transcription networks}, that motivate our
study.  We then frame the evolutionary question in the language of computational
learning theory, summarize our contributions and discuss related work.

\subsection{Representation in Biology}

Biological systems appear to function successfully with greatly restricted
representation classes. The nature of circuits found in biological systems may
vary, but some aspects -- such as \emph{sparsity} -- are common.  Specifically,
the interacting components in many biological circuits are sparsely connected.
Biological circuits are often represented as networks or graphs, where the
vertices correspond to entities such as neurons or molecules and the edges to
connections or interactions between pairs of entities. For example, both neural
networks~\cite{Watts:1998} and networks of metabolic reactions in the
cell~\cite{Wagner:2001,Barabasi:2000} have been described by ``small-world''
models, where a few ``hub'' nodes have many edges but most nodes have few edges
(and consequently, the corresponding graphs have small diameter).  An associated
property observed in biological networks is \emph{modularity}: a larger network
of interacting entities is composed of smaller modules of (functionally related)
entities~\cite{Hartwell:1999}.  Both the ``small-world'' description and
modularity of biological networks are consistent with the more general theme of
sparsity.

\begin{figure}[!t]
\centering
\subfigure[~]{
\includegraphics[width=0.47\textwidth]{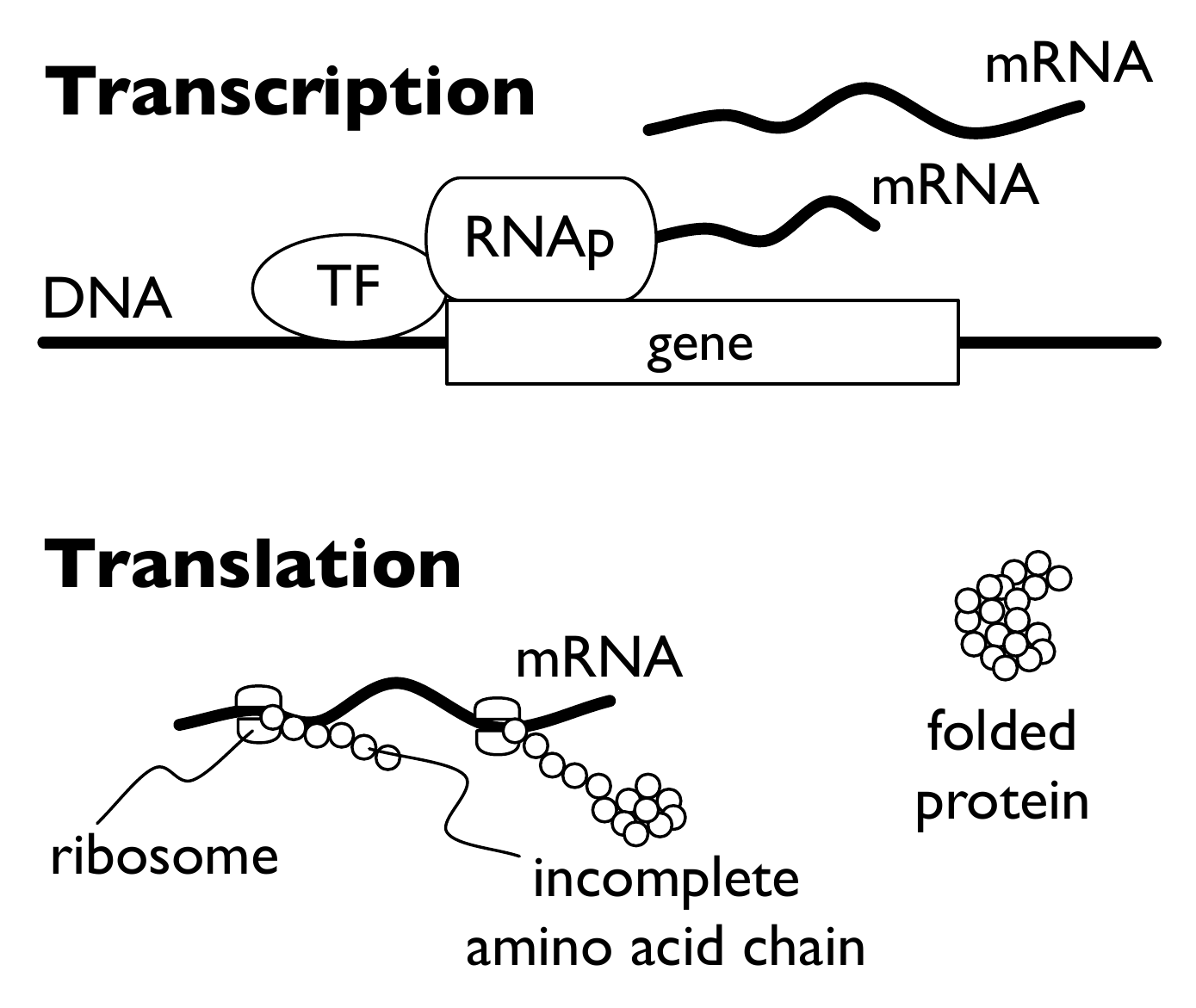}
\label{fig:biology}
}
\subfigure[~]{
\includegraphics[width=0.47\textwidth]{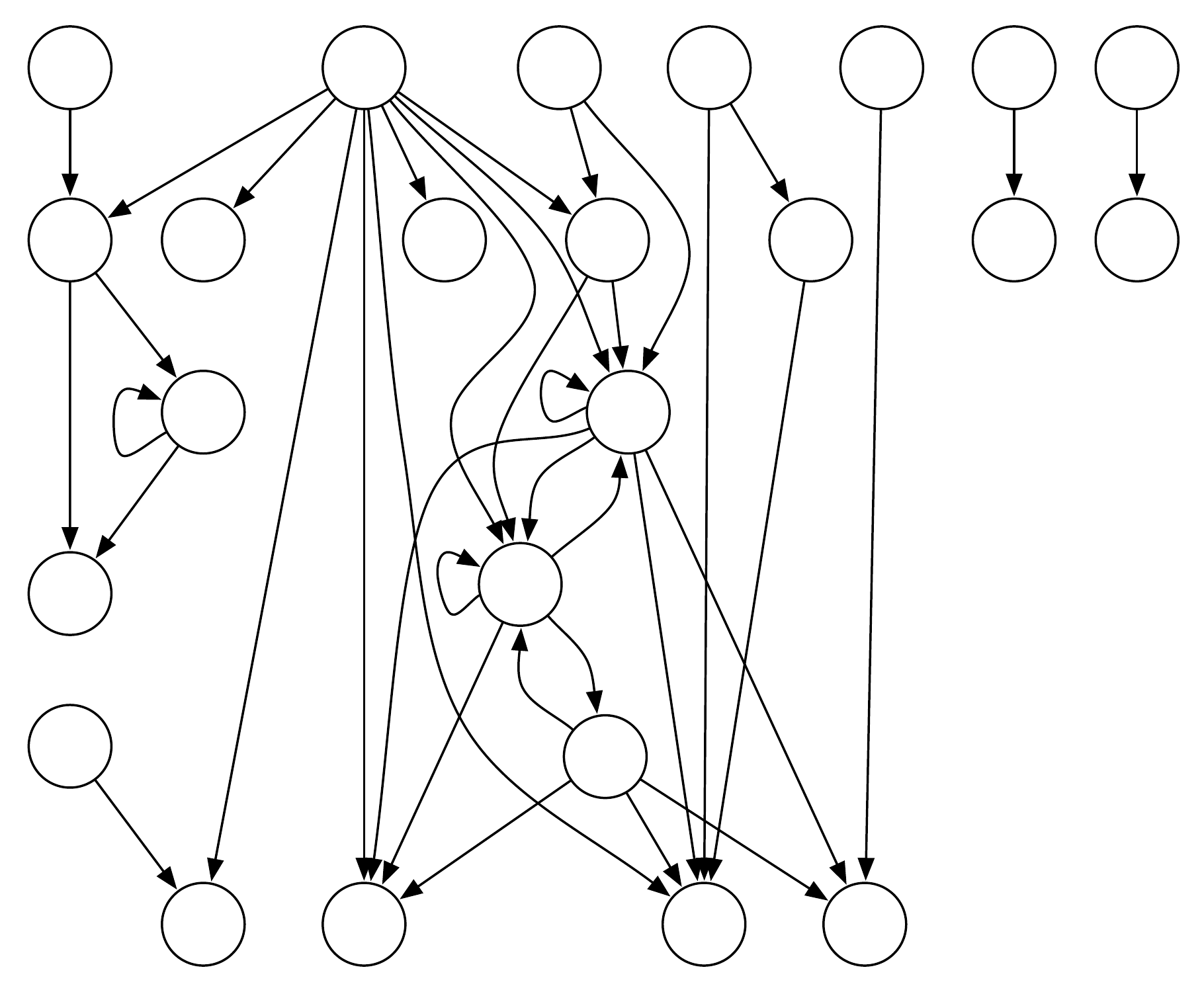}
\label{fig:network}
}
\caption{(a)~Schematic of transcription (top) and translation (bottom).
Here, a transcription factor (TF) binds to DNA close to a gene in a way that
increases gene expression by encouraging RNA polymerase (RNAp) to transcribe
the gene and so produce mRNA.  The mRNA is then translated by ribosomes to
produce sequences of amino acids that ultimately fold into proteins.
Only a small number of transcription factors
directly regulate any gene. Note that a transcription factor's action can also
decrease gene expression. For a more complete picture, see \eg~\cite{Alon:2006}.
(b)~Topology of the transcription network of respiration and redox reactions in
yeast. $X \rightarrow Y$ represents that transcription factor $X$ regulates the
expression of $Y$. Note that this real network has cycles.
Adapted from~\cite{Murray:2011}.}
\end{figure}

We focus on transcription networks, which are a specific class of networks of
interacting genes and proteins that are involved in the production of new
protein. Alon provides an accessible and mathematical introduction to
transcription networks and other biological circuits~\cite{Alon:2006}; below and
in Figure~\ref{fig:biology}, we present a simplified account that motivates this
work. Genes are \emph{transcribed} to produce mRNA, which is then
\emph{translated} into sequences of amino acids that ultimately fold into
proteins.\footnote{In reality, this is a dynamical system where the rates of
production are important. Note that this process need not be linear: a gene (mRNA
transcript) can be transcribed (translated) multiple times, not only in series
but also in parallel fashion.  We also ignore
other \emph{epigenetic} effects, \ie molecular modifications to DNA that do not
change its sequence but alter gene expression,~\eg the addition of methyl groups
to nucleotides in a way that physically blocks transcription.}
In a transcription network, a gene's transcription may be regulated by a set of
proteins called \emph{transcription factors}.
These transcription factors may increase or decrease a gene's transcription by
physically binding to regions of DNA that are typically close to the gene.
In natural systems, only a small number of transcription factors
regulate any single gene, and so transcription networks are sparsely connected.
For example, Balaji~\etal studied a yeast
transcription network of 157 transcription factors regulating 4,410 genes. They
observed this network to have 12,873 interactions (edges) where each gene was
regulated on average by about 2.9 transcription factors, the distribution of
in-degrees was well-described by an exponential fit, and only about 45 genes had
an in-degree of 15 or greater~\cite{Balaji:2006}.

The number of transcription factors varies from hundreds in a bacterium to
thousands in a human cell. Some transcription factors are always present in the
cell and can be thought of as representing a \emph{snapshot} of the
environment~\cite{Alon:2006}.
For example, the presence of sugar molecules in the environment may cause
specific transcription factors to be \emph{activated}, enabling them to regulate
the production of other proteins.  One of these proteins could be an
\emph{end-product}, such as an enzyme that catalyzes a metabolic reaction
involving the sugar. Alternatively, the transcription factor could regulate
another transcription factor that itself
regulates other genes -- we view this as intermediate computation -- and may
participate in further ``computation'' to produce the desired end-result.

While transcription networks may include cycles (loops), here for simplicity we
focus on systems that are directed acyclic graphs, and the resulting computation
can be viewed as a circuit. We illustrate a small, real transcription network in
Figure~\ref{fig:network}. These circuits are by necessity shallow due to
a temporal constraint, that the time required for sufficient quantities of
protein to be produced is of the same order of magnitude as cell-division
time.\footnote{Other kinds of networks, such as signaling networks, operate by
changing the shapes of proteins. The fact that these transformations are rapid
may allow for much larger depth. Note that fast conformational changes govern
how transcription factors directly process information from the environment in
order to regulate gene expression.  In our example, a sugar molecule binds to a
transcription factor and changes its shape in a way that alters its ability to
bind to DNA.} For example, Luscombe~\etal measured the shortest path length (in
number of intermediate nodes) between transcription factors and regulated genes
corresponding to terminal nodes (leaves) in a yeast transcription network. In
the static network, the mean such path length was 4.7 and the longest path
involved 12 intermediate transcription factors~\cite{Luscombe:2004}.

\subsection{Our Contributions}

First, our contribution is conceptual. We believe that the study of evolvability
from a computational standpoint will benefit by understanding the representation
complexity required to evolve a certain concept class. Motivated by the previous
discussion, in the case of transcription networks, it appears essential that the
representation used be a constant depth and fan-in (boolean or arithmetic)
circuit. Of course, any function that can be represented by such a circuit can
depend only on a constant number of input variables. We ask the
question, when we restrict attention to functions in a given class that depend
only on a constant number of variables, when can evolution succeed?

Second, we show that the class of sparse linear functions, those that depend
only on a constant number of variables, under a large class of smooth
distributions, can be evolved using sparse linear functions as representations,
when the performance is measured using squared error. The number of variables
used by the representations is larger than the number of variables in the
\emph{ideal function} and depends on the \emph{smoothness} parameter of the
distribution. According to our notion of $\Delta$-smooth $G$-nice distributions
(Defn.~\ref{defn:afghanistan}), the density function of a smooth distribution
is obtained by convolution of an arbitrary density with a product measure on
$[-\sqrt{3}\Delta, \sqrt{3}\Delta]^n$ (alternatively, drawing a point from the
smooth distribution is equivalent to drawing a point from an arbitrary
distribution and adding a (noise) vector from a product distribution). 

A linear function is represented by a weighted arithmetic circuit with only one
addition gate (alternatively, by a depth-two circuit with a layer of
multiplication gates and some constant inputs).\footnote{There is a natural
tradeoff between fan-in and depth, that may be useful, depending on which is the
more severe constraint.} Also, the number of generations required for evolution
to succeed depends polynomially on the sparsity $k$ of the target linear
function, the smoothness parameter $\Delta$ of the distribution and the inverse
of the target accuracy $\epsilon$, and has no dependence on the dimension $n$ of
the input space. The number of mutations explored at each generation depends
polynomially in $n$ and $1/\epsilon$. Thus, our result shows
\emph{attribute-efficient} evolvability of sparse linear functions, in the sense
of Littlestone~\cite{Littlestone:1988}. For the precise statement, see
Theorem~\ref{thm:sparse_linear} in Section~\ref{sec:sparse_linear}.

Valiant also proposed a stronger selection mechanism -- when natural selection
aggressively selects the (almost) best mutation, rather than merely a beneficial
one -- called evolution by optimization. Our second result requires a much
stronger distributional assumption -- the correlation $\corr(x_i, x_j) \leq
1/(2k)$ -- where $k$ is the sparsity of the target  linear function (see
Defn.~\ref{defn:bhutan}). Under such distributions, we show that under evolution
by optimization, sparse linear functions can be evolved by representations with
the same sparsity. The mechanism we propose and its analysis is inspired by the
greedy orthogonal matching pursuit algorithms in signal
processing~\cite{Donoho:2006-recovery,Tropp:2004-greed}. Unlike the previous
evolutionary algorithm, this one requires initialization, \ie the evolutionary
process begins with the $0$ function. As in the previous case, the number of
generations required depends polynomially on the sparsity $k$ of the target
linear function, the inverse of the accuracy parameter $\epsilon$, but has no
dependence on the total number of attributes $n$. The precise statement appears
as Theorem~\ref{thm:greedy} in Section~\ref{sec:greedy}.

\subsubsection*{Related Work}

The question of proper vs. improper learning has been studied in computational
learning theory. A separation between the two kinds is known, unless $\NP =
\RP$. However, most interesting PAC-learnable classes can be learned using
thresholds of low-degree polynomials, and do not seem to require the full
generality of polynomial-sized circuits.\footnote{For example, the classes of
$k$-CNF, $k$-term DNF, decision lists and low-rank decision trees, can all be
represented as PTFs.} In this context, Valiant's disjunction algorithm under the
uniform distribution~\cite{Valiant:2009-evolvability}, Kanade {\it et al.}'s
algorithm for homogeneous half-spaces under radially symmetric
distributions~\cite{KVV:2010-drift}, and P. Valiant's algorithm for linear
(polynomial) functions using squared loss~\cite{Valiant:2012-real}, are
\emph{proper} evolutionary mechanisms, \ie the representation used is from the
same class as the ideal function.  In the first two cases, it is straightforward
to show that if the target depends only on a constant number of variables, the
evolutionary mechanism also succeeds using representations that depend only on a
constant number of variables. Thus, attribute-efficient evolution can be
achieved.

The problem of learning sparse linear functions has been studied under various
names in several fields for many applications, \eg recovering sparse solutions
to (underdetermined) linear systems of equations~\cite{Donoho:2009-sparse}, or
recovering sparse representations with a redundant
dictionary~\cite{Mallat:2008,Elad:2010}; compressive sampling or compressed
sensing for sparse signal reconstruction~\cite{Candes:2008}; optimization with
regularization or sparsity-inducing penalties in machine
learning~\cite{Bach:2012}; sparse coding for learning an overcomplete
basis~\cite{Olshausen:1997}, or for denoising in image and video
processing~\cite{Elad:2010}. This area is too vast to review here; Bruckstein
\emph{et al.} have an excellent survey~\cite{Donoho:2009-sparse}.  Learning the
sparsest linear function is equivalent to finding the sparsest solution to a
system of linear equations (assuming there is no noise in the data). In general,
this problem is $\NP$-hard and the currently best-known approximation factor
depends on the norm of the pseudo-inverse of the matrix~\cite{Natarajan:1995}.
Thus, some assumption on the distribution seems necessary. Our evolution based
on optimization algorithm (Section~\ref{sec:greedy}) is essentially the greedy
orthogonal matching pursuit algorithm of Tropp~\cite{Tropp:2004-greed} and
Donoho \emph{et al.}~\cite{Donoho:2006-recovery}, cast in the language of
evolvability; these algorithms are also known in statistical modeling as forward
stepwise regression~\cite{Daniel:1999,Hastie:2001}.

Finally, the question of \emph{attribute-efficient} regression in the PAC (or
SQ) model is a natural one. Here, the goal would be to design a polynomial time
algorithm for producing an $\epsilon$-accurate linear function, with sample
complexity that is polynomial in the sparsity $k$ of the target function and the
inverse of the target accuracy $\epsilon$, and only polylogarithmic in $n$, the
total number of attributes. Under mild boundedness assumptions on the
distribution, this can be achieved by setting up an $L_1$-regularized
optimization problem; the output classifier may not be sparse in light of the
$\NP$-hardness result mentioned above. We note that under the distributional
assumption made in this paper, finding the \emph{sparsest} linear function that
fits the data is also easy in the PAC/SQ setting, since the solution to the
optimization problem in this case is unique. The focus in our work is different,
namely showing that simple evolutionary mechanisms can succeed, while using
representations that are themselves sparse linear functions at all times.

\subsubsection*{Organization}

In Section~\ref{sec:notation}, we give an overview of Valiant's evolution model
and describe the concept classes and class of distributions considered in this
paper.  Section~\ref{sec:algorithms} contains the mechanisms for evolving sparse
linear functions. We conclude in Section~\ref{sec:conclusion} with some
discussion and directions for future work.

\section{Model and Preliminaries}
\label{sec:notation}
We first provide an overview of the evolvability framework of
Valiant~\cite{Valiant:2009-evolvability}. The description here differs slightly
from Valiant's original formulation and includes some subsequent extensions (for
more details the reader is referred to
\cite{Valiant:2009-evolvability,Feldman:2008-evolvability,
Feldman:2009-robustness, Valiant:2012-real, Kanade:2012-thesis}).

\subsection{Valiant's Evolvability Framework}
\label{sec:notation-model}

Let $X$ denote a set of instances, \eg $X = \reals^n$ or $X = \{0, 1\}^n$. We
assume that the representation length of each $x \in X$ is captured by the
parameter $n$. To avoid excessive notation, we will keep this size parameter
implicit in our description of the model. Let $D$ be a distribution over $X$.
Each $x \in X$ can be thought of as the description of an environmental setting,
the inputs to any circuit of an organism. $D$ denotes the distribution over the
possible environmental settings an organism may experience in a lifetime. Let $f
: X \rightarrow Y$ (typically $Y = \reals$ or $Y = \{0, 1\}$) denote the
\emph{ideal function}, the best behavior in each possible environmental
setting.

\subsubsection*{Representations}

A creature is a string representation that encodes an efficiently computable
function $r : X \rightarrow Y$, \ie there is an efficient Turing Machine that,
given the description string $\langle r \rangle$ and $x \in X$, outputs $r(x)$.  

In this work, our focus is characterizing different evolutionary mechanisms
based on the complexity of representations used. 
The complexity of a representation is measured by the function it computes.  Let
$H : X \rightarrow Y$ be a class of functions. For $R \subseteq \{0, 1\}^*$,  we
say that $R$ \emph{represents} $H$, if there is a map, $\sigma : R \rightarrow
H$, and if there exists an \emph{efficient} Turing machine that, given input $r
\in R$ and $x \in X$, outputs $(\sigma(r))(x)$. Henceforth, by abuse of notation
we will use $r$ to denote both the representation and the function it computes,
$\sigma(r)$. 

\subsubsection*{Evolutionary Algorithms}

The performance of a representation $r$ is measured using a loss function $\ell
: Y \times Y \rightarrow \reals^+$, such that $\ell(y, y) = 0$. For a function
$g : X \rightarrow Y$, define the expected loss with respect to the ideal
function $f : X \rightarrow Y$, under distribution $D$, as $\loss_{f, D}(g) =
\E_{x \sim D}[\ell(g(x), f(x))]$.\footnote{This definition does not require the
expected loss to be bounded, but we will mainly be interested in situations when
that is the case.} The goal of evolution is to reach some representation $r^*$
such that $\loss_{f, D}(r^*) < \epsilon$. In the following discussion, we use
the notation: $f$ the ideal function, $\epsilon$ the target accuracy, $D$ the
target distribution over $X$ and $\loss_{f, D}(g)$ the expected loss function.  \medskip \\
\noindent{\bf Mutator}: A mutator $\Mut(r, \epsilon)$, for a set of representations $R$, is
a polynomial-time randomized Turing machine that takes as input a representation
$r \in R$ and accuracy parameter $\epsilon$ and outputs a multiset $\Neigh(r,
\epsilon) \subseteq R$. The running time requirement on $\Mut$ also ensures that
$|\Neigh(r, \epsilon)|$ is polynomially bounded. \medskip \\
\noindent{\bf Selection}: (Natural) Selection is based on the empirical performance
of each representation. Let $s : R \times [0, 1] \rightarrow \naturals$ be a
sample size function. First, the mutation algorithm, $\Mut(r, \epsilon)$, is run
to produce multiset $\Neigh(r, \epsilon)$. Then, an i.i.d. sample $\langle x^i
\rangle_{i=1}^s$ is drawn from the distribution $D$ over $X$, where $s = s(r,
\epsilon)$.  Denote the empirical performance of each $r^\prime \in \Neigh(r,
\epsilon) \cup \{r \}$ as
\[ \hat{\loss}_{f, D}(r^\prime) = \frac{1}{s}\sum_{i=1}^s \ell(r^\prime(x^i),
f(x^i)) \]
Finally, let $t : R \times [0, 1] \rightarrow \reals$ be a tolerance function.
Two possible selection mechanisms are considered.
\begin{enumerate}
\item {\bf Selection based on beneficial and neutral mutations} ($\bnsel$): Let 
\[ \Bene = \{r^\prime \in \Neigh(r, \epsilon) ~|~ \hat{\loss}_{f, D}(r^\prime) \leq
\hat{\loss}_{f, D}(r) - t(r, \epsilon) \} \]  
denote the set of beneficial mutations and let 
\[ \Neut = \{r^\prime \in \Neigh(r, \epsilon) ~|~ |\hat{\loss}_{f, D}(r^\prime) -
\hat{\loss}_{f, D}(r)| <  t(r, \epsilon) \} \]
denote the neutral mutations, with respect to tolerance function $t$. Both
$\Bene$ and $\Neut$ are treated as multisets (the multiplicity of any
representation is the same as that in $\Neigh(r, \epsilon)$). Selection
operates as follows: if $\Bene \neq \emptyset$, $r^\prime$ is randomly selected
from $\Bene$ as the surviving creature at the next generation.  If $\Bene =
\emptyset$ and $\Neut \neq\emptyset$, then $r^\prime$ is selected randomly from
$\Neut$ as the surviving creature at the next generation.  Otherwise, $\bot$ is
produced signifying failure of evolution.
\item {\bf Selection based on optimization} ($\optsel$): Let $\widehat{\opt} =
\displaystyle\min_{r^\prime \in \Neigh(r, \epsilon)} \hat{\loss}_{f,
D}(r^\prime)$.  If $\widehat{\opt} > \hat{\loss}_{f, D}(r) + t(r, \epsilon)$,
then $\bot$ is produced signifying failure of evolution.  Otherwise, consider
the multiset, $\best = \{ r^\prime \in \Neigh(r, \epsilon) ~|~ \hat{\loss}_{f,
D}(r^\prime) \leq \widehat{\opt} + t(r, \epsilon) \}$, and then $r^\prime$ is
chosen from $\best$ randomly as the surviving creature at the next generation.
\end{enumerate}

\noindent Thus, while the selection rule $\bnsel$ only chooses some beneficial
(or at least neutral) mutation, $\optsel$ aggressively picks the (almost) best
mutation from the available pool. \medskip

We denote by $r^\prime \leftarrow \Sel[R, \Mut, s, t](r, \epsilon)$ the fact
that $r^\prime$ is the surviving creature in the next generation after one
mutation and selection operation on the representation $r$ and accuracy
parameter $\epsilon$. Here, $\Sel$ may be one of the two selection rules
described above. For $\Sel$ to be feasible we require that the size function $s$
is polynomially bounded (in $n$ and $1/\epsilon$) and that the inverse of the tolerance
function $t$ is polynomially sandwiched, \ie there exists polynomials $p_1(n,
1/\epsilon)$ and $p_2(n, 1/\epsilon)$ such that $1/p_1(n, 1/\epsilon) \leq t(r,
\epsilon) \leq 1/p_2(n, 1/\epsilon)$ for every $r \in R$ and $\epsilon > 0$.
\medskip \\
\noindent {\bf Evolutionary Algorithm}: An evolutionary algorithm $\evalg$ is a
tuple $(R, \Mut, s, t, \Sel)$. When $\evalg$ is run starting from $r_0 \in R$
with respect to distribution $D$ over $X$, ideal function $f : X \rightarrow Y$,
loss function $\ell$ and parameter $\epsilon$, a sequence $r_0, r_1, r_2,
\ldots$ is produced, where $r_i \leftarrow \Sel[R, \Mut, s, t](r_{i - 1},
\epsilon)$. If $r_i = \bot$ for some $i$, we consider evolution as halted and
$r_j = \bot$ for $j > i$. We say that $\evalg$ succeeds at generation $g$, if
$g$ is the smallest index for which the expected loss $\loss_{f, D}(r_g) \leq
\epsilon$.

\begin{definition}[Evolvability \cite{Valiant:2009-evolvability}] We say that a
concept class $C$ is evolvable with respect to loss function $\ell$ and selection
rule $\Sel$, under a class of distributions $\Dists$ using a representation
class $H$, if there exists a representation scheme $R \subseteq \{0, 1\}^*$,
such that $R$ represents $H$, and there exists an evolutionary algorithm $\evalg
= (R, \Mut, s, t, \Sel)$, such that for every $D \in \Dists$, every $f \in C$,
every $\epsilon > 0$, and every $r_0 \in R$, with probability at least $1 -
\epsilon$, $\evalg$ run starting from $r_0$ with respect to $f, D, \ell,
\epsilon$, produces $r_g$ for which $\loss_{f, D}(r_g) < \epsilon$.
Furthermore, the number of generations $g$ required for evolution to succeed
should be bounded by a polynomial in $n$ and $1/\epsilon$.  \end{definition}

\begin{remark} If the evolutionary algorithm succeeds only for a specific
starting representation $r_0$, we say $C$ is evolvable with
\emph{initialization}. \end{remark}

\begin{remark} If the functions in concept class $C$ depend only on $k$
variables, we say the evolutionary algorithm is attribute-efficient, if the size
function, $s$, is polylogarithmic in $n$, and polynomial in $k$ and
$1/\epsilon$, and the number of generations, $g$, is polynomial in $k$ and
$1/\epsilon$, but does not depend on $n$.
\end{remark}

The definition presented above varies slightly from the definition of Valiant,
in the sense that we explicitly focus on the complexity of representations used
by the evolutionary algorithm. As discussed in the introduction, we focus on
concept classes where each function depends on \emph{few} (constant) input
variables.\footnote{These functions have been referred to as juntas in the
theory literature. We avoid using this nomenclature as we restrict our attention
to specific functional forms, such as linear functions, with $k$ relevant
variables.} 


\subsection{Sparse Linear Functions} 
\label{sec:notation-class}

Our main result in this paper concerns the class of sparse linear functions.  We
represent a linear function from $\reals^n \rightarrow \reals$ by a vector $w
\in \reals^n$, where $x \mapsto w \cdot x$.  For a vector $w \in \reals^n$,
$\lznorm{w}$ is the number of non-zero elements of $w$.

For any $0 \leq l < u$ and integer $k$, define the class of linear functions:
\[
\lin^k_{l, u} = \{ x \mapsto w \cdot x ~|~ \lznorm{w} \leq k, \forall i,
w_i = 0 \mbox{ or } l \leq |w_i| \leq u \}
\]
Thus, $\lin^k_{l, u}$ is the class of $k$-sparse linear functions, where the
``influence'' of each variable is upper and lower bounded.\footnote{We do not
use the word ``influence'' in the precise technical sense here.}
\todoea{Why ``influence'' rather than simply magnitude?}

Let $D$ be a distribution over $\reals^n$. For $w, w^\prime \in \reals^n$, define the inner
product $\ip{w}{w^\prime} = \E_{x \sim D}[(w \cdot x) (w^\prime \cdot x)]$,
where $w \cdot x = \sum_{i = 1}^n w_i x_i$ denotes the standard dot product in
$\reals^n$. In this paper, we use $\ltwonorm{w}$ to denote $\sqrt{\ip{w}{w}}$
(and not $\sqrt{\sum_{i} w_i^2}$). To avoid confusion, whenever necessary, we
will refer to the quantity $\sqrt{\sum_{i} w_i^2}$ explicitly if we mean the
standard Euclidean norm. 
%
%

\subsubsection*{Distribution Classes}

We use two classes of distributions for our results in this paper. We define
them formally here. \medskip 

\noindent{\bf Smooth Bounded Distributions}: We consider the class of smooth
bounded distributions over $\reals^n$. The concept of smoothed analysis of
algorithms was introduced by Spielman and Teng~\cite{ST:2004} and recently the
idea has been used in learning theory~\cite{KST:2009,KKM:2013}. We consider
distributions that are bounded and have $0$ mean. Formally,
distributions we consider are defined as:

\begin{definition}[$\Delta$-Smooth $G$-Nice Distribution]
\label{defn:afghanistan} A distribution $D$ is a $\Delta$-smooth $G$-nice
distribution if it is obtained as follows. Let $\tilde{D}$ be some distribution
over $\reals^n$, and let $U^n_a$ denote the uniform distribution over $[-a,
a]^n$. Then $D = \tilde{D} * U^n_{\sqrt{3}\Delta}$ is obtained by the convolution of
$\tilde{D}$ with $U^n_{\sqrt{3} \Delta}$.\footnote{We could perform convolution
with a spherical Gaussian distribution, however, this would make the resulting
distribution unbounded. All results in this paper hold if we work with
sub-Gaussian distributions and consider convolution with a spherical Gaussian
distribution with variance $\Delta^2$.  In this case, we would be required to use
Chebychev's inequality rather than Hoeffding's bound to show that the empirical
estimate is close to the expected loss with high probability.} Furthermore, $D$
satisfies the following:
\begin{enumerate}
\item $\E_D[x] = 0$
\item For all $i$, $\E_D[x_i^2] \leq 1$
\item For every $x$ in the support of $D$, $\sum_{i = 1}^n x_i^2 \leq G^2$
\end{enumerate}
\end{definition}

\noindent{\bf Incoherent Distributions}: We also consider \emph{incoherent}
distributions.\footnote{This terminology is adapted from incoherence of
matrices, \eg see~\cite{Donoho:2009-sparse}.} For a distribution $D$ over
$\reals^n$, the coherence is defined as $\max_{i, j} \corr(x_i, x_j)$, where
$\corr(x_i, x_j)$ is the correlation between $x_i$ and $x_j$. Again, we consider
bounded distributions with zero mean. We also require the variance to be upper
and lower bounded in each dimension. Formally, the distributions we consider are
defined as:

\begin{definition}[$\mu$-Incoherent $(\Delta, G)$-Nice Distribution]
\label{defn:bhutan} A distribution $D$ is a $\mu$-incoherent $(\Delta, G)$-nice
distribution if the following hold:
\begin{enumerate}
\item $\E_D[x] = 0$
\item For all $i$, $\Delta^2 \leq \E_D[x_i^2] \leq 1$
\item For all $i$, $j$, $\max_{i, j} \corr(x_i, x_j) \leq \mu$
\item For all $x$ in the support of $D$, $\sum_{i=1}^n x_i^2 \leq G^2$
\end{enumerate}
\end{definition}


We say a linear function represented by $w \in \reals^n$ is $W$-bounded if
$\sum_{i=1}^n w_i^2 \leq W^2$. We use the notation $w(x) = w \cdot x$. Suppose
$f, w$ are $W$-bounded linear functions, and distribution $D$ is such that for every
$x$ in the support of $D$, $\sum_{i=1}^n x_i^2 \leq G^2$.  We consider the
squared loss function, which for $y, y^\prime \in \reals$ is $\ell(y^\prime, y) =
(y^\prime - y)^2$.  Then, for any $x$ in the support of $D$, $\ell(f(x), w(x))
\leq 4 W^2G^2$. Thus, standard Hoeffding bounds imply that
if $\langle x^i \rangle_{i=1}^s$ is an i.i.d. sample drawn from $D$, then
\begin{align}
\Pr\left[\left| \frac{1}{s} \sum_{i=1}^s \ell(w(x^i), f(x^i)) - \loss_{f, D}(w)
\right| \geq \tau \right] &\leq 2 \exp\left( -\frac{ s \tau^2}{8W^2G^2}\right)
\label{eqn:concentration}
\end{align}


Finally, for linear functions $w$ ($x \mapsto w \cdot x$), let $\NZ(w) = \{ i
~|~ w_i \neq 0 \}$ denote the non-zero variables in $w$, so $\lznorm{w} =
|\NZ(w)|$. Then, we have the following Lemma. The proof appears in
Appendix~\ref{app:notation-class}.

\begin{lemma} \label{lemma:amsterdam} Let $D$ be a $\Delta$-smooth $G$-nice
distribution (Defn.~\ref{defn:afghanistan}), let $w \in \reals^n$ be a vector
and consider the corresponding linear function, $x \mapsto w \cdot x$. Then the
following are true:
\begin{enumerate}
\item For any $1 \leq i \leq n$, $w_i^2 \leq \frac{\ip{w}{w}}{\Delta^2}$.
\item There exists an $i$ such that $w_i^2 \leq
\frac{\ip{w}{w}}{|\NZ(w)|\Delta^2}$.
\end{enumerate}
\end{lemma}

\section{Evolving Sparse Linear Functions}
\label{sec:algorithms}
In this section, we describe two evolutionary algorithms for evolving sparse
linear functions. The first evolves the class $\lin^k_{l, u}$ under the class of
$\Delta$-smooth $G$-nice distributions (Defn.~\ref{defn:afghanistan}), using the
selection rule $\bnsel$. The second evolves the class $\lin^k_{0, u}$ under the
more restricted class of $(1/2k)$-incoherent $(\Delta, G)$-nice distributions
(Defn.~\ref{defn:bhutan}), using the selection rule $\optsel$. We
first define the notation used in the rest of this section.\smallskip \\

\noindent{\bf Notation}: $D$ denotes the target distribution over $X =
\reals^n$, $f$ denotes the ideal (target) function. The inner product
$\ip{\cdot}{\cdot}$ and $2$-norm $\ltwonorm{\cdot}$ of functions are with
respect to the distribution $D$. $[n]$ denotes the set $\{1, \ldots, n\}$. For
$S \subseteq [n]$, $f^S$ denotes the best linear approximation of $f$ using the
variables in the set $S$; formally,
\begin{align}
f^S = \underset{w \in \reals^n~:~ w_i = 0 ~\vee~i \in S}\argmin \ltwonorm{f -
w}^2 \nonumber 
\end{align}
Finally, recall that for $w \in \reals^n$, $\NZ(w) = \{i ~|~ w_i \neq 0 \}$ and
$\sparsity(w) = |\NZ(w)|$. A vector $w$ represents a linear function, $x \mapsto
w \cdot x$. The vector $e^i$ has $1$ in coordinate $i$ and $0$ elsewhere and
corresponds to the linear function $x \mapsto x_i$. Thus, in this notation,
$\corr(x_i, x_j) = \ip{e^i}{e^j}/(\ltwonorm{e^i}\ltwonorm{e^j})$. The accuracy
parameter is denoted by $\epsilon$. 

\subsection{Evolving Sparse Linear Functions Using $\bnsel$}
\label{sec:sparse_linear}

We present a simple mechanism that evolves the class of sparse linear functions
$\lin^k_{l, u}$ with respect to $\Delta$-smooth $G$-nice distributions (see
Defn.~\ref{defn:afghanistan}). The representation class also consists of
sparse linear functions, but with a greater number of non-zero entries than the
\emph{ideal function}. We also assume that a linear function is represented by
$w \in \reals^n$, where each $w_i$ is a real number. (Handling the issues of
finite precision is standard and is avoided in favor of simplicity.) Define the
parameters $K = 5184(k/\Delta)^4(u/l)^2$ and $B = 10 uk /\Delta$. Formally, the
representation class is:
\[ 
R = \{ w ~|~ \sparsity(w) \leq K, w_i \in [-B, B] \}
\]
The important point to note is that the parameters $K$ and $B$ do not depend on
$n$, the total number of variables.

Next, we define the mutator. Recall that the mutator is a randomized algorithm
that takes as input an element $r \in R$ and accuracy parameter $\epsilon$, and
outputs a multiset $\Neigh(r, \epsilon) \subseteq R$. Here, $\Neigh(r,
\epsilon)$ is populated by $m$ independent draws from the following procedure,
where $m$ will be specified later (see the proof of Theorem~\ref{thm:sparse_linear}).
Starting from $w \in R$, define the mutated representation $w^\prime$, output by
the mutator, as:
\begin{enumerate}
\item {\em Scaling}: With probability $1/3$, choose $\gamma \in [-1, 1]$ uniformly at
random and let $w^\prime = \gamma w$. 
\item {\em Adjusting}: With probability $1/3$, do the following.  Pick $i \in
\NZ(w) = \{ i~|~ w_i \neq 0 \}$ uniformly at random. Let $w^\prime$ denote the
mutated representation, where $w^\prime_j = w_j$ for $j \neq i$, and choose
$w^\prime_i \in [-B, B]$ uniformly at random.
\item With the remaining $1/3$ probability, do the following:
\begin{enumerate}
\item {\em Swapping}: If $|\NZ(w)| = K$, choose $i_1 \in \NZ(w)$ uniformly at random.
Then, choose $i_2 \in [n] \setminus \NZ(w)$ uniformly at random. Let $w^\prime$
be the mutated representation, where $w_j^\prime = w_j$ for $j \neq i_1, i_2$.
Set $w_{i_1}^\prime = 0$ and choose $w_{i_2}^\prime \in [-B, B]$ uniformly at
random. In this case, $\sparsity(w^\prime) = \sparsity(w) = K$
with probability $1$, and hence $w^\prime \in R$.
\item {\em Adding}: If $|\NZ(w)| < K$, choose $i \in [n] \setminus \NZ(w)$ uniformly
at random. Let $w^\prime$ be the mutated representation, where $w_j^\prime =
w_j$ for $j \neq i$, and choose $w^\prime_i \in [-B, B]$ uniformly at random.
\end{enumerate}
\end{enumerate}

Recall that $f \in \lin^k_{l, u}$ denotes the ideal (target) function and $D$ is the
underlying distribution that is $\Delta$-smooth $G$-nice (see
Defn.~\ref{defn:afghanistan}). Since we are working with the squared loss
metric, $\ell(y^\prime, y) = (y^\prime - y)^2$, the expected loss for any $w \in
R$ is given by $\loss_{f, D}(w) = \ltwonorm{f - w}^2 = \ip{f - w}{f - w}$.  We
will show that for any $w \in R$, if $\ltwonorm{f - w}^2 > \epsilon$, with
non-negligible (inverse polynomial) probability, the above procedure produces a
mutation $w^\prime$ that decreases the expected loss by at least some inverse
polynomial amount. Thus, by setting the size of the neighborhood $m$ large
enough, we can guarantee that with high probability there will always exist a
beneficial mutation.
%

To simplify notation, let $S = \NZ(w)$. Recall that $f^S$ denotes the best
approximation to $f$ using variables in the set $S$; thus, $\ltwonorm{f - w}^2 =
\ltwonorm{f - f^S}^2 + \ltwonorm{f^S - w}^2$. At a high level, the argument for
proving the success of our evolutionary mechanism is as follows: If
$\ltwonorm{f^S - w}^2$ is large, then a mutation of the type ``scaling'' or
``adjusting'' will get $w$ closer to $f^S$, reducing the expected loss. (The
role of ``scaling'' mutations is primarily to ensure that the representations
remain bounded.) If $\ltwonorm{f^S - w}^2$ is small and $S \neq \NZ(f)$, there
must be a variable in $\NZ(f) \setminus S$, that when added to $w$ (possibly by
swapping), reduces the expected loss. Thus, as long as the representation is far
from the evolutionary target, a \emph{beneficial} mutation is produced with high
probability.

More formally, let $w^\prime$ denote a random mutation produced as a result of
the procedure described above.  We will establish the desired result by proving
the following claims.
\begin{claim} \label{claim:apple} If $\ltwonorm{w} \geq 2 \ltwonorm{f^S}$, then
with probability at least $1/12$, $\loss_{f, D}(w^\prime) \leq \loss_{f, D}(w) -
\ltwonorm{f^S - w}^2/12$. In particular, a ``scaling'' type mutation achieves
this. \end{claim}
\begin{claim} \label{claim:banana} When $\ltwonorm{w} \leq 2 \ltwonorm{f^S}$,
then with probability at least $\Delta \ltwonorm{f^S - w}/(6K^2 B)$, $\loss_{f,
D}(w^\prime) \leq \loss_{f, D}(w) - 3 \Delta^2\ltwonorm{f^S - w}^2/(4|S|^2)$. In
particular, an ``adjusting'' type mutation achieves this. \end{claim}
\begin{claim} \label{claim:cantaloupe} When $\ltwonorm{f^S - w} \leq
l^2\Delta^2/(4KB)$, but $\NZ(f) \not\subseteq S$, then with probability at least
$\Delta \ltwonorm{f - w}/(6KBnk)$, $\loss_{f, D}(w^\prime) \leq \loss_{f, D}(w)
- \Delta^2 \ltwonorm{f-w}^2/(16k^2)$. In particular, a mutation of type
``swapping'' or ``adding'' achieves this.
\end{claim}

\noindent Note that when $\NZ(f) \subseteq S$, then $f^S = f$. Thus, in this case
when $\loss_{f, D}(w) = \ltwonorm{f^S - w}^2 \leq \epsilon$, the evolutionary
algorithm has succeeded. \medskip 

The proofs of the above Claims are provided in
Appendix~\ref{app:sparse_linear}. We now prove our main result using the
above claims.

\begin{theorem} Let $\Dists$ be the class of $\Delta$-smooth $G$-nice
distributions over $\reals^n$ (Defn.~\ref{defn:afghanistan}). Then the
class $\lin^k_{l, u}$ is evolvable with respect to $\Dists$, using the
representation class $\lin^K_{0, B}$, where $K = O((k/\Delta)^4 (u/l)^2)$ and $B
= O(uk/\Delta)$, using the mutation algorithm described in this section, and the
selection rule $\bnsel$.  Furthermore, the following are true:
\begin{enumerate}
\item The number of generations required is polynomial in $(u/l)$, $1/\epsilon$,
$1/\Delta$, and is independent of $n$, the total number of attributes. 
\item The size function $s$, the number of points used to calculate empirical
losses, depends polylogarithmically on $n$, and
polynomially on the remaining parameters. 
\end{enumerate}
\label{thm:sparse_linear} \end{theorem}
\begin{proof}
The mutator is as described in this section. Let 
\[ p = \min\left\{\frac{1}{12}, \frac{l^2\Delta^3}{24K^3B^2}, \frac{\Delta
\sqrt{\epsilon}}{6KBnk}\right\},\] 
and let 
\[ \alpha = \min\left\{\frac{l^4 \Delta^4}{192 K^2B^2}, \frac{3 l^4
\Delta^6}{64K^4B^2}, \frac{\epsilon\Delta^2}{16k^2}\right\}.\] 
Now, by Claims~\ref{claim:apple}, \ref{claim:banana} and \ref{claim:cantaloupe},
if $\ltwonorm{f - w}^2 \geq \epsilon$, then the mutator outputs a mutation that
decreases the squared loss by $\alpha$ with probability at least $p$.

Recall that $K = 5184 (k/\Delta)^4 (u/l)^2$ and $B = 10uk/\Delta$.  Now, let $g
= 20 K G^2 B^2/\alpha$ (recall that $G^2$ is the bound on $\sum_{i} x_i^2$ for
$x$ in the support of the distribution). We will show that evolution succeeds in
at most $g$ generations. Note that $g$ has no dependence on $n$, the number of
attributes, and polynomial dependence on the remaining parameters. Define $m =
p^{-1} \ln(2g /\epsilon)$, and at each time step we have that $|\Neigh(w,
\epsilon)| = m$.  Note that together with the observation above, this implies
that except with probability $\epsilon/2$, for $1 \leq i \leq g$, if $w^i$ is
the representation at time step $i$, $\Neigh(w^i, \epsilon)$ contains a mutation
that decreases the loss by at least $\alpha$, if $\loss_{f, D}(w^i) \geq
\epsilon$. 

Now, let $t = 3\alpha / 5$ be the \emph{tolerance function}, set $\tau =
\alpha/5$ and let $s = (200 g KG^2B^2/\alpha^2) \ln(4m/\epsilon)$ be the
\emph{size function}. Note that $\sum_{i} w_i^2 \leq KB^2$ for $w \in R$ (this
also holds for $f$, since $k < K$ and $u < B$). If $\langle x^i \rangle_{i=1}^s$
is an i.i.d. sample drawn from $D$, for each $\bar{w}$ of the $mg$
representations that may be considered in the neighborhoods for the first $g$
time steps, using~(\ref{eqn:concentration}), it holds that $|\hat{\loss}_{f,
D}(\bar{w}) - \loss_{f, D}(\bar{w})| \leq \tau$ simultaneously except with
probability $\epsilon/2$ (by a union bound). Thus, allowing for failure
probability $\epsilon$, we assume that we are in the case when the neighborhood
always has a mutation that decreases the expected loss by $\alpha$ (whenever the
expected loss of the current representation is at least $\epsilon$) and that all
empirical expected losses are $\tau$-close to the true expected losses.

Now let $w$ be the representation at some generation such that $\loss_{f, D}(w)
\geq \epsilon$, let $w^\prime \in \Neigh(w, \epsilon)$ such that $\loss_{f,
D}(w^\prime) \leq \loss_{f, D}(w) - \alpha$. Then, it is the case that
$\hat{\loss}_{f, D}(w^\prime) \leq \hat{\loss}_{f, D}(w) - 3\alpha/5$ (when
$\tau = \alpha/5$). Hence, for tolerance function $t = 3 \alpha/5$, for the
selection rule using $\bnsel$, $w^\prime \in \Bene$. Consequently $\Bene \neq
\emptyset$. Hence, the representation at the next generation is chosen from
$\Bene$. Let $\tilde{w}$ be the chosen representation. It must be the case that
$\hat{\loss}_{f, D}(\tilde{w}) \leq \hat{\loss}_{f, D}(w) - t$. Thus, we have
$\loss_{f, D}(\tilde{w}) \leq \loss_{f, D}(w) - \alpha/5$. Hence, the expected
loss decreases at least by $\alpha/5$.

Note that at no point can the expected loss be greater than $4KG^2B^2$ for any
representation in $R$. Hence, in at most $20 KG^2B^2/\alpha$ generations,
evolution reaches a representation with expected loss at most $\epsilon$. Note
the only parameter introduced which has an inverse polynomial dependence on $n$
is $p$. This implies that $s$ only has polylogarithmic dependence on $n$. This
concludes the proof of the theorem.
\end{proof}

\begin{remark} We note that the same evolutionary mechanism works when evolving
the class $\lin^k_{0, u}$, as long as the sparsity $K$ of the representation
class is allowed polynomial dependence on $1/\epsilon$, the inverse of the accuracy parameter.
This is consistent with the notion of attribute-efficiency, where the goal is
that the information complexity should be polylogarithmic in the number of
attributes, but may depend polynomially on $1/\epsilon$.
\end{remark}

\subsection{Evolving Sparse Linear Functions Using $\optsel$}
\label{sec:greedy}
In this section, we present a different evolutionary mechanism for evolving
sparse linear functions. This algorithm essentially is an adaptation of a greedy
algorithm commonly known as orthogonal matching pursuit (OMP) in the signal
processing literature (see ~\cite{Donoho:2006-recovery, Tropp:2004-greed}). Our
analysis requires stronger properties on the distribution: we show that
$k$-sparse linear functions can be evolved with respect to $1/(2k)$-incoherent
$(\Delta, G)$-nice distributions (Defn.~\ref{defn:bhutan}). Here, the
selection rule used is selection using \emph{optimization}
($\optsel$).\footnote{Valiant showed that selection using optimization was
equivalent to selection using beneficial and neutral
mutations~\cite{Valiant:2009-evolvability}. However, this reduction uses
representation classes that may be somewhat complex. For restricted
representation classes, it is not clear that such a reduction holds. In
particular, the necessary ingredient seems to be \emph{polynomial-size}
memory.} Also, the algorithm is guaranteed to succeed only with
\emph{initialization} from the $0$ function.  Nevertheless, this evolutionary
algorithm is appealing due to its simplicity and because it never uses a
representation that is not a $k$-sparse linear function.

Recall that $f \in  \lin^k_{0, u}$ is the ideal (target)
function.\footnote{Here, we no longer need the fact that each coefficient in the
target linear function has magnitude at least $l$.} Let 
\[ R = \{w ~|~ \sparsity(w) \leq k, w_i \in [-B, B] \}, \]
where $B = 10uk/\Delta$. Now, starting from $w \in R$, define the action of the
mutator as follows (we will define the parameters $\lambda$ and $m$ later in the
proof of Theorem~\ref{thm:greedy}):
\begin{enumerate}
\item {\em Adding}: With probability $\lambda$, do the following. Recall that
$\NZ(w)$ denotes the non-zero entries of $w$. If $|\NZ(w)| < k$, choose $i \in [n]
\setminus \NZ(w)$ uniformly at random. Let $w^\prime$ be such that $w^\prime_j =
w_i$ for $j \neq i$, and choose $w^\prime_i \in [-B, B]$ uniformly at random. If
$\NZ(w)= k$, let $w^\prime = w$. Then, the multiset $\Neigh(w, \epsilon)$ is
populated by $m$ independent draws from the procedure just described.
\item With probability $1 - \lambda$, do the following:
\begin{enumerate}
\item {\em Identical}: With probability $1/2$, output $w^\prime = w$.
\item {\em Scaling}: With probability $1/4$, choose $\gamma \in [-1, 1]$ uniformly at
random and let $w^\prime = \gamma w$.
\item {\em Adjusting}: With probability $1/4$, do the following.  Pick $i \in \NZ(w)$
uniformly at random.  Let $w^\prime$ be such that $w^\prime_j = w_j$ for $j \neq
i$, and choose $w^\prime_i \in [-B, B]$ uniformly at random.
\end{enumerate}
Then, the multiset $\Neigh(w, \epsilon)$ is populated by $m$ independent draws
from the procedure just described.
\end{enumerate}

One thing to note in the above definition is that the mutations produced by the
mutator at any given time are correlated, \ie they are all either of the kind
that add a new variable, or all of the kind that just manipulate existing
variables.  At a high level, we prove the success of this mechanism as follows:
\begin{enumerate}
\item Using mutations of type ``scaling'' or ``adjusting,'' a representation
that is close to the \emph{best} in the space, \ie $f^S$, is evolved.
\item When the representation is (close to) the best possible using current
variables, adding one of the variables that is present in the \emph{ideal
function}, but not in the current representation, results in the greatest
reduction of expected loss. Thus, selection based on optimization would always
add a variable in $\NZ(f)$. By tuning $\lambda$ appropriately, it is ensured
that with high probability, candidate mutations that add new variables are not
chosen until evolution has had time to approach the \emph{best} representation
using existing variables.
\end{enumerate}

To complete the proof we establish the following claims.

\begin{claim} \label{claim:date} If $\ltwonorm{f^S - w} \leq
\sqrt{\epsilon}/{2k}$, then if $S \subsetneq \NZ(f)$, there exists $i \in \NZ(f)
\setminus S$ and $-B < a < b < B$, such that for any $\gamma \in [a, b]$,
$\loss_{f, D}(w + \gamma e^i) \leq \loss_{f, D}(w) - \epsilon/(4k^2)$ and for
any $j \not\in \NZ(f)$, $\beta \in [-B, B]$, $\loss_{f, D}(w + \beta e^j) \geq
\loss_{f, D}(w + \gamma e^i) + \epsilon/(4k^3)$. Furthermore, $b - a \geq
\sqrt{(k+1) \epsilon}/k^2$. \end{claim}

\begin{claim} \label{claim:elderberry} Conditioned on the mutator not outputting
mutations that add a new variable, with probability at least $\min\{1/16,
\ltwonorm{f^S - w}/(16k^2B)\}$, there exists a mutation that reduces the squared
loss by at least $\ltwonorm{f^S - w}^2/(12k^2)$. \end{claim}

The proofs of Claims~\ref{claim:date} and \ref{claim:elderberry} are not
difficult and are provided in Appendix~\ref{app:greedy}. Based on the above
claims we can prove the following theorem:

\begin{theorem} \label{thm:greedy} Let $\Dists$ be the class of
$1/(2k)$-incoherent $(\Delta, G)$-nice distributions over $\reals^n$
(Defn.~\ref{defn:bhutan}).
Then, the class $\lin^k_{0, u}$ is evolvable with respect to $\Dists$ by an
evolutionary algorithm, using the mutation algorithm described in this section,
selection rule $\optsel$, and the representation class $R = \lin^k_{0, B}$,
where $B = 10 uk/\Delta$.  Furthermore, the following are true:
\begin{enumerate}
\item The number of generations $g$ is polynomial in $1/\epsilon$, $k$,
$1/\Delta$, but independent of the dimension~$n$.
\item The size function $s$, the number of points used to calculate the
empirical losses, depends polylogarithmically on $n$ and polynomially on the
remaining parameters.
\end{enumerate}
\end{theorem}
\begin{proof} The proof is straightforward, although a bit heavy on notation; we
provide a sketch here. The mutator is as described in this section. Let
\[ p = \min \left\{ \frac{1}{16}, \frac{\sqrt{\epsilon}}{64 k^3 B},
\frac{\sqrt{(k+1)\epsilon}}{k^2} \right\}, \]
and let
\[ \alpha = \min \left\{ \frac{\epsilon}{4k^3}, \frac{\epsilon}{192k^4} \right\}
= \frac{\epsilon}{192k^4}. \]
Also, let $\tau = \alpha/5$ and let $t = 3\alpha/5$ be the \emph{tolerance
function}.

First, we show that between the ``rare'' time steps when the mutator outputs
mutations that add a new variable, evolution has enough time to \emph{stabilize}
(reach close to local optimal) using existing variables. To see this, consider a
sequence of coin tosses, where the probability of heads is $\lambda$ and the
probability of tails is $1 - \lambda$. Let $Y_i$ be the number of tails between
the $(i-1)\th$ and $i\th$ heads. Except with probability $\epsilon/(4(k+1))$,
$Y_i > \epsilon/(4(k+1)\lambda)$ by a simple union bound.  Also, by Markov's
inequality, except with probability $\epsilon/(4(k+1))$, $Y_i <
4(k+1)/(\epsilon\lambda)$. Thus, except with probability $\epsilon/2$, we have
$\epsilon/(4(k+1)\lambda) \leq Y_i \leq 4(k+1)/(\epsilon\lambda)$ for $i = 1, 2,
\ldots, k+1$. Let $g = 4(k+1)^2/(\epsilon\lambda) + (k+1)$. This ensures that,
except with probability $\epsilon/2$, after $g$ time steps, at least $(k+1)$
time steps where the mutator outputs mutations of type ``adding'' have occurred,
and the first $k$ of these occurrences are all separated by at least
$\epsilon/(4(k+1) \lambda)$ time steps of other types of mutations.

Also, let $m = p^{-1}\ln(4g/\epsilon)$ and let $s = (200gkG^2B^2/\alpha^2)
\ln(4m/\epsilon)$ be the \emph{size function}.  These values ensure that for
$g$ generations, except with probability $\epsilon/2$, the mutator always
produces a mutation that had probability at least $p$ of being produced
(conditioned on the type of mutations output by the mutator at that time step),
and that for all the representations concerned, $|\hat{\loss}_{f, D}(w) -
\loss_{f, D}(w)| \leq \tau$, where $\tau = \alpha/5$. Thus, allowing the process
to fail with probability $\epsilon$, we assume that none of the undesirable
events have occurred.

We will show that the steps with mutations other than ``adding'' are sufficient
to ensure that evolution reaches the (almost) best possible target with the
variables available to it. In particular, if the set of available variables is
$S$, the representation $w$ reached by evolution will satisfy $\ltwonorm{f^S -
w}^2 \leq \epsilon/(2k^2)$. For now, suppose that this is the case. 

We claim by induction that evolution never adds a ``wrong'' variable, \ie one
that is not present in the target function $f$. The base case is trivially true,
since the starting representation is $0$. Now suppose, just before a ``heads''
step, the representation is $w$, such that $S = \NZ(w)$ and $\ltwonorm{f^S - w}
\leq \epsilon/(2k^2)$. The current step is assumed to be a ``heads'' step, thus
the mutator has produced mutations by adding a new variable. Then, using
Claim~\ref{claim:date}, we know that there is a mutation $w^\prime$ in
$\Neigh(w, \epsilon)$ such that $\loss_{f, D}(w^\prime) < \loss_{f, D}(w) -
\epsilon/(4k^2)$ (obtained by adding a correct variable). Since $\alpha <
\epsilon/(4k^2)$ and $\tau = \alpha/5$, it must be the case that
$\hat{\loss}_{f, D}(w^\prime) \leq \hat{\loss}_{f, D}(w) - 3 \alpha/5$. This
ensures that the set $\best$, for selection rule $\optsel$ is not empty.
Furthermore, we claim that no mutation that adds an \emph{irrelevant} variable
can be in $\best$. Suppose $w^{\prime\prime}$ is a mutation that adds an
\emph{irrelevant} variable; according to Claim~\ref{claim:elderberry},
$\loss_{f, D}(w^{\prime\prime})> \loss_{f, D}(w^\prime) + \alpha$, and hence
$\hat{\loss}_{f, D}(w^{\prime\prime}) > \hat{\loss}_{f, D}(w^\prime) + t$. This ensures
that every representation in $\best$ corresponds to a mutation that adds some
\emph{relevant} variable. Thus, the evolutionary algorithm never adds any
\emph{irrelevant} variable.

Finally, note that during a ``tails'' step (when the mutator produces mutations of types
other than ``adding''), as long as $\ltwonorm{f^S - w}^2 \geq \epsilon/(4k^2)$,
there exists a mutation that reduces the expected loss by at least
$\epsilon^2/(192k^4) = \alpha$. This implies that the set $\best$ is non-empty
and for the values of tolerance $t = 3\alpha/5$ and $\tau = \alpha/5$, any
mutation from the set $\best$ reduces the expected loss by at least $\alpha/5$.
(This argument is identical to the one in Theorem~\ref{thm:sparse_linear}.)
Since the maximum loss is at most $4kB^2G^2$ for the class of distributions and
a representation $w$ from the set $R$; in at most $20kB^2G^2/\alpha$ steps, a
representation satisfying $\loss_{f, D}(w) \leq \epsilon/(4k^2)$ must be
reached. Note that once such a representation is reached, it is ensured that the
loss does not increase substantially, since with probability at least $1/2$, the
mutator outputs the same representation. Hence, it is guaranteed that there is
always a neutral mutation. Thus, before the next ``heads'' step, it must be the
case that $\ltwonorm{f^S - w}^2 \leq \epsilon/2k^2$. If $\lambda$ is set to
$\epsilon\alpha/(80 k(k+1)B^2G^2)$, the evolutionary algorithm using the
selection rule $\optsel$ succeeds. 

It is readily verified that the values of $g$ and $s$ satisfy the claims in the
statement of the theorem.
\end{proof}

\section{Conclusion and Future Work}
\label{sec:conclusion}
In this work, we provided simple evolutionary mechanisms for evolving sparse
linear functions, under a large class of distributions. These evolutionary
algorithms have the desirable properties that the representations used are
themselves sparse linear functions, and that they are attribute-efficient in the
sense that the number of generations required for evolution to succeed is
independent of the total number of attributes.

Strong negative results are known for distribution-independent evolvability of
boolean functions, \eg even the class of conjunctions is not
evolvable~\cite{Feldman:2011-LTF}. However, along the lines of this work, it is
interesting to study whether under restricted classes of distributions,
evolution is possible for simple concept classes, using representations of low
complexity. Currently, even under (biased) product distributions, no
evolutionary mechanism is known for the class of disjunctions, except via
Feldman's general reduction from CSQ algorithms. Even if the queries made by the
CSQ algorithm are simple, Feldman's reduction uses intermediate representations
that randomly combine queries made by the algorithm, making the representations
quite complex.

A natural extension of our current results would be to study fixed-degree sparse
polynomials. Another interesting direction is to study circuits with sigmoidal
or other non-linear filters on the gates, which arise naturally in molecular
systems.  The suitable class of boolean functions to study is low-weight
threshold functions, which includes disjunctions and conjunctions. The class of
smooth bounded distributions may be an appropriate starting place for studying
evolvability of these classes. For example, is the class of low-weight threshold
functions evolvable under smooth distributions, or at least log-concave
distributions?

\subsection*{Acknowledgments} 
We would like to thank Leslie Valiant for helpful discussions and comments on an
earlier version of this paper. We are grateful to Frank Solomon for discussing
biological aspects related to this work.

\bibliography{all-refs}
\bibliographystyle{plain}

\newpage
\appendix 
\section{Omitted Proofs}

\subsection{Proofs from Section~\ref{sec:notation-class}}
\label{app:notation-class}
\begin{proof}[Proof of Lemma~\ref{lemma:amsterdam}] Note that for any $x \sim
D$, we can write $x = \tilde{x} + \eta$, where $\tilde{x}$ is drawn from some
smooth bounded distribution, and $\eta$ is drawn from the uniform distribution
over $[-\sqrt{3} \Delta, \sqrt{3} \Delta]^n$. Note that $\eta$ and $\tilde{x}$
are independent, and all components of $\eta$ are independent. First, we observe
that $\E[x_i^2] = \E[(\tilde{x}_i + \eta_i)^2] \geq \E[\eta_i^2] = \Delta^2$.
Now, consider the following:
\begin{align*}
\ip{w}{w} &= \E\left[\left(\sum_{i=1}^n w_i x_i\right)^2\right] \\
&= \E\left[\left(\sum_{i = 1}^n w_i (\tilde{x}_i + \eta_i) \right)^2 \right] \\ 
&= \E\left[\left(\sum_{i=1}^n w_i \tilde{x}_i\right)^2\right] +
2\E\left[\left(\sum_{i=1}^n w_i \tilde{x}_i \right)\left(\sum_{i=1}^n w_i
\eta_i\right) \right] + \E\left[\left(\sum_{i=1}^n w_i \eta_i\right)^2\right] \\
&\geq \sum_{i=1}^n w_i^2 \E[\eta_i^2] \mbox{~~~~~~~Since $\eta_i$ are all independent}
\\
&\geq \sum_{i = 1}^n w_i^2 \Delta^2  \mbox{~~~~~~~~~~Since $\E[\eta_i^2] = \Delta^2$ by
definition} \\
&= \sum_{i \in \NZ(w)} w_i^2 \Delta^2
\end{align*}
The conclusions of the Lemma follow easily by looking at the above expression.
\end{proof}

\subsection{Proofs from Section~\ref{sec:sparse_linear}}
\label{app:sparse_linear}
\begin{proof}[Proof of Claim \ref{claim:apple}] We show that in this case, a
``scaling'' mutation achieves the desired result. Restricted to the direction
$w$, the best approximation to $f^S$ is $\frac{\ip{f^S}{w}}{\ltwonorm{w}^2}w$.
We have that
\[
\left\Vert \frac{\ip{f^S}{w}}{\ltwonorm{w}^2} w \right\Vert \leq
\ltwonorm{f^S} \leq \frac{\ltwonorm{w}}{2}
\]
Hence, if $\ip{f^S}{w} > 0$, for $\gamma \in [1/4, 3/4]$ (and similarly if
$\ip{f^S}{w} < 0$ for $\gamma \in [-3/4, -1/4]$ ), we have that
\begin{align*}
\ltwonorm{f^S - \gamma w}^2 &= \ltwonorm{f^S - w}^2 + 2 (1 - \gamma) \ip{f^S -
w}{w} + (1 - \gamma)^2 \ltwonorm{w}^2 \\
&\leq \ltwonorm{w - f^S}^2 - (1 - \gamma) \ltwonorm{w}^2 + (1 - \gamma)^2
\ltwonorm{w}^2 \\
&= \ltwonorm{w - f^S}^2 - (\gamma - \gamma^2) \ltwonorm{w}^2
\intertext{Finally, by observing that for $\gamma \in [1/4, 3/4]$, $\gamma -
\gamma^2 \geq 3/16$ and that by the triangle inequality, $\ltwonorm{w} \geq
(2/3)\ltwonorm{f^S - w}$ when $\ltwonorm{w} \geq 2 \ltwonorm{f^S}$, we obtain}
\ltwonorm{f^S - \gamma w}^2 &\leq \ltwonorm{f^S - w}^2 - \frac{1}{12}
\ltwonorm{f^S - w}^2
\end{align*}
We note that $\loss_{f, D}(w^\prime) = \ltwonorm{f - f^S}^2 + \ltwonorm{f^S -
\gamma w}^2$ and $\loss_{f, D}(w) = \ltwonorm{f - f^S}^2 + \ltwonorm{f^S -
w}^2$. An appropriate value of $\gamma$ is chosen with probability at least
$1/4$, and combined with the probability of choosing a scaling mutation we get
the desired result.
\end{proof}

\begin{proof}[Proof of Claim~\ref{claim:banana}] Here, we appeal to a mutation
that adjusts the relative weights of the variables within the set $S = \NZ(w)$.
Consider the vector $f^S - w$, and note that $\NZ(f^S -w ) \subseteq S$. Let
$r^S = f^S - w$ denote the residual, which lies in the space spanned by $S$.
Now consider
\begin{align*}
\ltwonorm{r_S}^2 = \ip{r^S}{r^S} &= \sum_{i \in S} r^S_i \ip{e^i}{r^S}
\intertext{Here, $e^i$ is the unit vector representing the linear function $x
\mapsto e^i \cdot x = x_i$. Therefore, there must exist an $i$ for which the
following is true:}
r^S_i\ip{e^i}{r^S} &\geq \frac{\ltwonorm{r^S}^2}{|S|}
\intertext{We appeal to Lemma~\ref{lemma:amsterdam} (part 1), which implies that
$|r^S_i| \leq \sqrt{\ip{r^S}{r^S}/\Delta^2} = \ltwonorm{r^S}/{\Delta}$, to
conclude that}
|\ip{e^i}{r^S}| &\geq \frac{\ltwonorm{r^S} \Delta}{|S|}
\end{align*}
Let $\beta = \frac{\ip{e^i}{r^S}}{\ltwonorm{e^i}^2}$ and suppose $w^\prime =
w + \gamma e^i$ for $\gamma \in [\beta - |\beta|/2, \beta +  |\beta|/2]$. We
then have
\begin{align*}
\ltwonorm{f^S - (w + \gamma e^i)}^2 &= \ltwonorm{f^S - w}^2 - 2 \gamma \ip{f^S -
w}{e^i} + \gamma^2 \ltwonorm{e^i}^2
\intertext{Recall that $f^S - w = r^S$ and note that $\ip{e^i}{r^S}$ and
$\gamma$ have the same sign. This, combined with the above equation, gives}
\ltwonorm{f^S - (w + \gamma e^i)}^2 &= \ltwonorm{f^S - w}^2 - (2
|\gamma||\beta| - \gamma^2) \ltwonorm{e^i}^2
\intertext{Finally, note that for $|\gamma| \in [|\beta|/2, 3|\beta|/2]$, $- 2
|\gamma| |\beta| + \gamma^2 \leq - 3 |\beta|^2/4$, hence, the
above equation and the fact that $\ltwonorm{e^i} \leq 1$ together yield}
\ltwonorm{f^S - (w + \gamma e^i)}^2 &\leq \ltwonorm{f^S - w}^2 - \frac{3}{4}
\beta^2 \ltwonorm{e^i}^2 \leq \ltwonorm{f^S - w}^2 - \frac{3}{4}
\frac{\ltwonorm{f^S - w}^2 \Delta^2}{|S|^2}
\end{align*}

Note that a suitable mutation $w^\prime = w + \gamma e^i$ is obtained with
probability at least $|\beta|/(6KB)$ ($1/3$ for choosing the right type of
mutation, $1/K$ for the correct choice of variable, and $|\beta|/(2B)$ for the
choosing the correct value of $w_i$).  Also note that
$|\beta| \geq \Delta \ltwonorm{f^S - w} / |S| \geq \Delta \ltwonorm{f^S - w} / K$.
For this to be a valid mutation, we also need to verify the fact that $w_i + \gamma
\in [-B, B]$, which is ensured by our choice of $B$. To see this, note that
$\ltwonorm{w} \leq 2 \ltwonorm{f^S} \leq 2 \ltwonorm{f}$ (the last part is
because $f^S$ is a projection of $f$ onto a lower dimensional space). Thus, by
Lemma~\ref{lemma:amsterdam} (part 1), $w_i \leq 2 \ltwonorm{f} / \Delta$. Also,
$|\beta| = |\ip{e^i}{r^S}|/\ltwonorm{e^i}^2 \leq \ltwonorm{r^S}/\ltwonorm{e^i}
= \ltwonorm{f^S - w}/\E[x_i^2] \leq 3 \ltwonorm{f}/\Delta$.
Thus, if $B > 13 \ltwonorm{f}/ (2\Delta)$, then $|w_i + \gamma| < B$ and the
mutation will be a valid one. Note that the maximum value of $\ltwonorm{f}$ for
$f \in \lin^k_{l, u}$ is $uk$. Thus, our choice of $B = 10 uk / \Delta$ is
sufficient.  This completes the proof of Claim~\ref{claim:banana}.
\end{proof}

\begin{proof}[Proof of Claim~\ref{claim:cantaloupe}] Finally, we show that if
$\ltwonorm{f^S - w}$ is very small, but $\NZ(f) \not\subseteq S$, then it must
be the case that a ``swapping'' or ``adding'' mutation is beneficial. We focus
on the swapping case, \ie when $|S| = K$; the adding step is a special case of
this. First, we observe that if there exists $i \in \NZ(f)$ such that $i \not\in
S$, then by using Lemma~\ref{lemma:amsterdam} (part 1), it must be the case that
$\ltwonorm{f - w}^2 \geq (f_i - w_i)^2 \Delta^2 = f_i^2 \Delta^2 \geq l^2 \Delta^2$.
Let $r = f - w$ denote the residual. Then, consider the following:
\begin{align}
\ip{r}{r} &=  \sum_{i \in \NZ(f) \setminus S} r_i \ip{e^i}{r} + \sum_{i \in S} r_i
\ip{e^i}{r} \nonumber
\intertext{Note that for all $i \in S$, $\ip{e^i}{f - f^S} = 0$, since $f^S$ is
the projection of $f$ onto the space spanned by the variables in $S$. Hence, if
$r^S = f^S - w$, the residual within the space spanned by $S$, then $r = f - f^S
+ r^S$. Thus, we have $\ip{e^i}{r} = \ip{e^i}{r^S} \leq \ltwonorm{r^S}$. Using
this we get, }
\ip{r}{r} &\leq \sum_{i \in \NZ(f) \setminus S} r_i \ip{e^i}{r} +
\ltwonorm{r^S} \sum_{i \in S} |r_i|  \nonumber
\intertext{Now, even by a very crude estimate, $|r_i| = |f_i - w_i| \leq 2 B$,
and hence by the condition in the statement of Claim~\ref{claim:cantaloupe},
that $\ltwonorm{r^S} =\ltwonorm{f^S - w} \leq l^2\Delta^2/(4KB)$, together with
the previous observation that $\ip{r}{r} = \ltwonorm{r}^2 =
\ltwonorm{f - w}^2 \geq l^2 \Delta^2$, we have that}
\frac{1}{2}\ip{r}{r} &\leq \sum_{i \in \NZ(f) \setminus S} r_i \ip{e^i}{r}
\nonumber
\intertext{We now appeal to Lemma~\ref{lemma:amsterdam} (part 1), which shows
that $|r_i| \leq \ltwonorm{r} / \Delta$, and
conclude that there exists an $i$ for which}
|\ip{e^i}{r}| &\geq \frac{\ltwonorm{r} \Delta}{2 k} \nonumber
\intertext{The crucial observation is that $|\NZ(f)| \leq k < K$. Let $\beta =
\frac{\ip{r}{e^i}}{\ltwonorm{e^i}^2}$. Finally, Lemma~\ref{lemma:amsterdam}
(part 2) implies that there exists an $i^\prime$ for which $w_{i^\prime}^2 \leq
\ltwonorm{w}^2/K$. We consider the mutation, $w^\prime = w + \gamma e^i -
w_{i^\prime} e^{i^\prime}$ for $\gamma \in [\beta - |\beta|/2, \beta +
|\beta|/2]$. Then, we have}
\ltwonorm{f - (w + \gamma e^i - w_{i^\prime} e^{i^\prime})}^2 &= \ltwonorm{f - (w
+ \gamma e^i)}^2 - 2 w_{i^\prime} \ip{f - (w + \gamma e^i)}{e^{i^\prime}} +
w_{i^\prime}^2 \ltwonorm{e^{i^\prime}}^2 \label{eqn:artichoke}
\intertext{We bound the first term on the right hand side of the above expression and then
the latter two.}
\ltwonorm{f - (w + \gamma e^i)}^2 &= \ltwonorm{r}^2 - 2 \gamma \ip{r}{e^i} +
\gamma^2 \ltwonorm{e^i}^2 \nonumber \\
&= \ltwonorm{r}^2 - (2 \gamma \beta  - \gamma^2) \ltwonorm{e^i}^2 \nonumber
\intertext{As in the proof of Claim~\ref{claim:banana}, for $|\gamma| \in
[|\beta|/2, 3|\beta|/2]$, we have that $- 2 \gamma \beta + \gamma^2 \leq -
3 \beta^2/4$. Hence,}
\ltwonorm{f - (w + \gamma e^i)}^2 &\leq \ltwonorm{r}^2 - \frac{3}{4} \beta^2
\label{eqn:broccoli}
\intertext{To bound the remaining two terms in (\ref{eqn:artichoke}), recall that
$f - w = f - f^S + r^S$ and that $\ip{f - f^S}{e^{i^\prime}} = 0$
(since $i^\prime \in S$). Thus, we get that}
- 2 w_{i^\prime} \ip{f - (w + \gamma e^i)}{e^{i^\prime}} + w_{i^\prime}^2
\ltwonorm{e^{i^\prime}}^2 &\leq 2 |w_{i^\prime}| |\ip{r^S + \gamma
e^i}{e^{i^\prime}}| + w_{i^\prime}^2 \ltwonorm{e^{i^\prime}}^2 \nonumber
\intertext{Using the fact that $\ltwonorm{r^S} \leq \gamma$, $|w_{i^\prime}|<
\gamma$ (which can be verified by the setting of $K$ below),
$|\ip{e^i}{e^{i^\prime}}| \leq 1$ and $\ltwonorm{e^{i^\prime}} \leq 1$, we
obtain}
- 2 w_{i^\prime} \ip{f - (w + \gamma e^i)}{e^{i^\prime}} + w_{i^\prime}^2
\ltwonorm{e^{i^\prime}}^2 &\leq 6 |w_{i^\prime}||\gamma| \label{eqn:cabbage}
\intertext{Recall that $|w_{i^\prime}| \leq \ltwonorm{w}/\sqrt{K}$. Also $\ltwonorm{w}
\leq \ltwonorm{f^S} + \ltwonorm{r^S} \leq 2 \ltwonorm{f^S} \leq 2 \ltwonorm{f}
\leq 2uk$, where we have used the fact that $\ltwonorm{r^S}$ is small. Combining
(\ref{eqn:artichoke}), (\ref{eqn:broccoli}),  (\ref{eqn:cabbage}), the fact that
$|w_i| \leq 2uk/\sqrt{K}$ and $|\gamma| \leq 3 |\beta|/2$, we get}
\ltwonorm{f - (w + \gamma e^i - w_{i^\prime} e^{i^\prime})}^2 &\leq
\ltwonorm{r}^2 - \frac{3}{4} \beta^2 + 18 |\beta|uk/\sqrt{K} \nonumber
\end{align}
Finally, we note that when $K > 5184 (k/\Delta)^4 (u/l)^2$, the above
equation ensures that the expected loss drops by at least $\beta^2/4$.  The probability of
choosing a swapping operations is $1/3$, of subsequently choosing the correct
pair is at least $1/(nK)$, and subsequently choosing the correct value of $w_i$
is at least $|\beta|/(2B)$. A simple calculation proves the statement of the
claim.
\end{proof}

\subsection{Proofs from Section~\ref{sec:greedy}}
\label{app:greedy}
\begin{proof}[Proof of Claim~\ref{claim:date}] Since $S \subseteq \NZ(f)$, the
residual $r = f - w$ is such that $\NZ(r) \subseteq \NZ(f)$. Consider $i \in \NZ(r)$
that maximizes $|r_i|\cdot \ltwonorm{e^i}$. Then, we have:
\begin{align}
\left|\frac{\ip{e^i}{r}}{\ltwonorm{e^i}} \right| &\geq
\frac{|r_i|\ip{e^i}{e^i}}{\ltwonorm{e^i}} - \sum_{j \in \NZ(r), j \neq i}
\frac{|r_j|\ip{e^i}{e^j}}{\ltwonorm{e^i}} \nonumber  \\
&\geq |r_i|\cdot \ltwonorm{e^i} - \frac{1}{2k} \sum_{j \in \NZ(r), j \neq i}
|r_j| \cdot \ltwonorm{e^j} \label{eqn:chives} \\
&\geq |r_i| \cdot \ltwonorm{e^i} \cdot \frac{k+1}{2k}, \nonumber
\end{align}
where in the last two steps we used the fact that $\corr(x_i, x_j) =
\ip{e^i}{e^j}/(\ltwonorm{e^i} \ltwonorm{e^j}) \leq 1/(2k)$ and that $|\NZ(r)
\setminus \{ i \}| \leq k-1$. On the other hand, for any $i^\prime \not\in
\NZ(r)$, we have
\begin{align}
\left| \frac{\ip{e^{i^\prime}}{r}}{\ltwonorm{e^{i^\prime}}} \right| &\leq
\sum_{j \in \NZ(r)} \frac{|r_j\ip{e^{i^\prime}}{e^j}|}{\ltwonorm{e^{i^\prime}}}
\nonumber \\
&\leq \frac{1}{2k} \sum_{j \in \NZ(r)} |r_j| \cdot \ltwonorm{e^j} \leq \frac{1}{2}
|r_i| \cdot \ltwonorm{e^i} \label{eqn:dill}
\end{align}

\noindent Here, again in the last two steps, we have used the fact that
$\corr(x_{i^\prime}, x_j) \leq 1/(2k)$ and that $|r_i| \cdot \ltwonorm{e^i}$ is
the largest such term.

First, we claim that if $\ltwonorm{r}^2 \geq \epsilon$, then the $i$ that
maximized $|r_i| \cdot \ltwonorm{e^i}$ must be from the set $\NZ(f) \setminus
S$. (Note that $\ltwonorm{r}^2 = \ltwonorm{f - w}^2 = \loss_{f, D}(w)$, so if
$\ltwonorm{r}^2 \leq \epsilon$, evolution has reached its goal.) By the
triangle inequality, $\sum_{i \in \NZ(r)} |r_i| \cdot \ltwonorm{e^i} \geq
\ltwonorm{r} \geq \sqrt{\epsilon}$. Hence, it must be the case that $|r_i| \cdot
\ltwonorm{e^i} \geq \sqrt{\epsilon}/{k}$. For contradiction, assume that $i \in
S$. Then, since $f^S$ is the projection of $f$ in the space spanned by $S$, we
have $\ip{e^i}{r} = \ip{e^i}{f^S - w}$, since $r = f - f^S + f^S - w$ and
$\ip{e^i}{f - f^S} = 0$. But, by the assumption of the claim, $|\ip{e^i}{r}|\leq
\ltwonorm{e^i} \cdot \ltwonorm{f^S - w} \leq \ltwonorm{e^i} \cdot
\sqrt{\epsilon}/(2k)$, and by (\ref{eqn:chives}), we know that $|\ip{e^i}{r}|
\geq \ltwonorm{e^i} \cdot (|r_i| \cdot \ltwonorm{e^i}) \cdot (k+1)/(2k) >
\ltwonorm{e^i} \cdot \ltwonorm{r} / (2k) \ge \ltwonorm{e^i} \cdot \sqrt{\epsilon}/(2k)$.
Thus, it cannot be the case that $i \in S$.

Let $w^\prime = w + \gamma e^i$. Then, 
\begin{align} 
\ltwonorm{f - (w + \gamma e^i)}^2 - \ltwonorm{f - w}^2 &= - 2 \gamma \ip{f -
w}{e^i} + \gamma^2 \ltwonorm{e^i}^2 \nonumber \\
&\leq - \ltwonorm{e^i}^2 \left(|\gamma| \cdot |r_i| \cdot \frac{k+1}{k} -
|\gamma|^2\right) \nonumber
\intertext{Now suppose $\gamma$ satisfies $1 - \delta \leq
(2|\gamma|k)/(|r_i|(k+1)) \leq 1 + \delta$, then using the fact that the quadratic
function on the RHS is maximized at $|\gamma| = (k+1) |r_i|/(2k)$, we have}
\ltwonorm{f - (w + \gamma e^i)}^2 - \ltwonorm{f - w}^2 &\leq - \ltwonorm{e^i}^2
\cdot \frac{r_i^2}{4} \cdot \frac{(k+1)^2}{k^2} \cdot (1 - \delta^2) \label{eqn:eggplant}
\end{align}

Note that for any $i^\prime \not\in S$, the ``best'' representation of the form
$w + \beta e^{i^\prime}$ is when $\beta =
\ip{e^{i^\prime}}{r}/\ltwonorm{e^{i^\prime}}^2$, and the corresponding reduction
in squared loss is $(\ip{e^{i^\prime}}{r})^2/\ltwonorm{e^{i^\prime}}^2$. Thus,
for any $i^\prime \neq i$, we have
\begin{align}
\ltwonorm{f - (w + \beta e^{i^\prime})}^2 - \ltwonorm{f - w}^2 &\leq -
\frac{\ip{e^{i^\prime}}{r}^2}{\ltwonorm{e^{i^\prime}}^2} \nonumber \\
&\leq - \frac{r_i^2}{4} \ltwonorm{e^i}^2 &\mbox{Using~(\ref{eqn:dill})}
\nonumber
\end{align}
Setting $\delta = \sqrt{1/(k+1)}$ completes the proof of the claim. To
see that $b - a \geq \sqrt{(k+1)\epsilon}/k^2$, notice that any $\gamma$,
such that $|\gamma| \in [(1 - \delta) ((k+1)/(2k)) |r_i|, (1 - \delta)
((k+1)/(2k)) |r_i|]$, achieves the claimed reduction in squared loss. Since
$|r_i| \cdot \ltwonorm{e^i} \geq \sqrt{\epsilon}/k$, we have that $|r_i| \geq
\sqrt{\epsilon}/k$. Hence, $b - a \geq 2 \delta ((k+1)/2k) (\sqrt{\epsilon}/k)
\geq \sqrt{(k+1) \epsilon}/k^2$, for $\delta = \sqrt{1/(k+1)}$.
\end{proof}

\begin{proof}[Proof of Claim~\ref{claim:elderberry}] 
The proof follows along the lines of the proofs of Claims~\ref{claim:apple} and
\ref{claim:banana}.  First, suppose that $\ltwonorm{w} \geq 2 \ltwonorm{f^S}$.
In this case, we claim that for $\gamma \in [1/2, 3/4]$, the mutation $\gamma w$
reduces the squared loss by at least $\ltwonorm{f^S - w}^2/12$. This analysis is
completely identical to that in Claim~\ref{claim:apple} and hence is omitted.
The only difference is that the probability that such a mutation is selected is
$1/16$, conditioned on the event that the mutator chooses mutations that don't
add an extra variable.

Next, we assume that $\ltwonorm{w} \leq 2 \ltwonorm{f^S}$. Let $r^S = f^S - w$.
Now, as in the proof of Claim~\ref{claim:date}, consider $i \in \NZ(r^S)$ (recall
that $\NZ(r^S) = S$) that maximizes $|r^S_i| \cdot \ltwonorm{e^i}$.
Then, the following is true:
\begin{align}
\left| \frac{\ip{e^i}{r^S}}{\ltwonorm{e^i}} \right| &\geq |r^S_i|
\frac{\ip{e^i}{e^i}}{\ltwonorm{e^i}} - \sum_{j \in S, j \neq i} |r^S_j|
\frac{\ip{e^i}{e^j}}{\ltwonorm{e^i}} \nonumber \\
&\geq |r^S_i| \cdot \ltwonorm{e^i} - \frac{1}{2k} \sum_{j \in S, j \neq i}
|r^S_j| \cdot \ltwonorm{e^j} \nonumber \\
&\geq |r^S_i| \cdot \ltwonorm{e^i} \cdot \frac{k+1}{2k} \label{eqn:fennel}
\end{align}
where in the last two steps we used the fact that $\corr(x_i, x_j) \leq 1/(2k)$
and that $|r^S_j| \cdot \ltwonorm{e^j}$ is maximized for $j = i$, and the fact
that $|S| \leq k$. Also, by the triangle inequality, we know that $\sum_{j \in
S} |r^S_j| \cdot \ltwonorm{e^j} \geq \ltwonorm{r^S}$; hence, by definition of
$i$, we have that $|r^S_i| \cdot \ltwonorm{e^i} \geq \ltwonorm{r^S}/k$. Now, let
$\beta = \ip{e^i}{r^S}/\ltwonorm{e^i}^2$, and for $\gamma \in [\beta -
|\beta|/2, \beta + |\beta|/2]$, consider the mutation $w + \gamma e^i$. We have,
\begin{align*}
\ltwonorm{f^S - (w + \gamma e^i)}^2 - \ltwonorm{f^S - w}^2 &= -2 \gamma
\ip{e^i}{r^S} + \gamma^2 \ltwonorm{e^i}^2 \\
&= - \ltwonorm{e^i}^2(2 |\gamma| |\beta| - |\gamma|^2) \\
&\leq - \frac{3}{4} \ltwonorm{e^i}^2 \beta^2 &\mbox{For $\gamma \in [\beta -
|\beta|/2, \beta + |\beta|/2]$} \\
&\leq - \frac{3}{16} \frac{\ltwonorm{r^S}^2}{k^2} &\mbox{Using
(\ref{eqn:fennel}) and the defn. of $\beta$}
\end{align*}

In order for $w + \gamma e^i$ to be a valid mutation, we need to check that
$|w_i| + 3 |\beta|/2 < B$. To see this, observe the following:
\begin{align*}
\ltwonorm{w}^2 &\geq \sum_{j \in \NZ(w)} w_j^2 \ltwonorm{e^j}^2 - \sum_{j_1 < j_2; j_1,
j_2 \in \NZ(w)} 2 |w_{j_1} w_{j_2}| |\ip{e^{j_1}}{e^{j_2}}| \\
&\geq \sum_{j \in \NZ(w)} w_j^2 \ltwonorm{e^j}^2 - \frac{1}{k} \sum_{j_1 < j_2}
|w_{j_1} w_{j_2}| \ltwonorm{e^{j_1}}\ltwonorm{e^{j_2}} \\
&\geq \frac{1}{2} \sum_{j \in \NZ(w)} w_j^2 \ltwonorm{e^j}^2 + \frac{1}{2k} \sum_{j_1 < j_2}
(|w_{j_1}| \ltwonorm{e^{j_1}} - |w_{j_2}| \ltwonorm{e^{j_2}})^2 \\
&\geq \frac{1}{2} \sum_{j \in \NZ(w)} w_j^2 \ltwonorm{e^j}^2 
\end{align*}

\noindent Hence, by a fairly loose analysis, $|w_i|  \leq 2 \ltwonorm{w}/\Delta
\leq 4 \ltwonorm{f}/\Delta \leq 4uk/\Delta$ (since $\ltwonorm{e^i} \geq \Delta$
for the class of distributions defined in Defn.~\ref{defn:bhutan}). Also
$|\beta| = |\ip{e^i}{r^S}|/\ltwonorm{e^i}^2 \leq \ltwonorm{r^S}/\ltwonorm{e^i}
\leq 3 \ltwonorm{f}/\Delta$ (since $\ltwonorm{r^S} \leq \ltwonorm{f^S} +
\ltwonorm{w} \leq 3 \ltwonorm{f^S} \leq 3 \ltwonorm{f}$). It's easy to see that
$|w_i| + (3/2) |\beta| \leq B$, for $B = 10 uk/\Delta$.

Finally, note that conditioned on the mutator choosing a mutation that doesn't
add new variables, the probability of choosing such a mutation is at least
$|\beta|/(8Bk)$ ($1/4$ for choosing a mutation of type ``adjusting'', $1/k$ for
choosing the appropriate variable to adjust and $|\beta|/(2B)$ for choosing the
correct value). Combining the claims for mutations of the ``scaling'' and
``adjusting'' types and taking the appropriate minimum values proves the
statement of the claim.
\end{proof}

\end{document}